\documentclass[conference,onecolumn]{IEEEtran}
\IEEEoverridecommandlockouts
\usepackage{cite}
\usepackage{amsmath,amssymb,amsfonts}
\usepackage{algorithm,algorithmic}
\usepackage{graphicx}
\usepackage{textcomp}
\usepackage{xcolor}
\def\BibTeX{{\rm B\kern-.05em{\sc i\kern-.025em b}\kern-.08em
    T\kern-.1667em\lower.7ex\hbox{E}\kern-.125emX}}

\usepackage{hyperref} 
\usepackage{tikz}
\usetikzlibrary{calc,shapes,positioning}
\usetikzlibrary{arrows}

\pdfoutput=1

\usepackage{url}
\usepackage{cite}
\usepackage{bm}

\usepackage{mystyle}

\usepackage{amsmath,graphicx}

\RequirePackage{algorithm}
\RequirePackage{algorithmic}

\hypersetup{  
  bookmarks=true,
  backref=true,
  pagebackref=false,
  colorlinks=true,
  linkcolor=blue,
  citecolor=red,
  urlcolor=blue,
  pdftitle={alphaBP},
  pdfauthor={Dong Liu},
  pdfsubject={}
}

\begin{document}
\title{$\alpha$ Belief Propagation as Fully Factorized Approximation}
\author{
  \IEEEauthorblockN{
    Dong Liu$^{1}$, Nima N. Moghadam$^{2}$, Lars~K. Rasmussen$^{1}$, Jinliang Huang$^{2}$, Saikat Chatterjee$^{1}$}

  \IEEEauthorblockA {$^{1}$ KTH Royal Institute of Technology, Stockholm, Sweden. }
  \IEEEauthorblockA {$^{2}$ Huawei Technologies Sweden AB, Stockholm, Sweden.}
  \IEEEauthorblockA{
    e-mail: \{doli, lkra, sach\}@kth.se, \{nima.najari.moghadam1, jinliang.huang\}@huawei.com}
}

\maketitle
\begin{abstract}
  Belief propagation (BP) can do exact inference in loop-free graphs, but its performance could be poor in graphs with loops, and the understanding of its solution is limited.
  This work gives an interpretable belief propagation rule that is actually minimization of a localized $\alpha$-divergence. We term this algorithm as $\alpha$ belief propagation ($\alpha$-BP). 
  The performance of $\alpha$-BP is tested in MAP (maximum a posterior) inference problems, where
  $\alpha$-BP can outperform (loopy) BP by a significant margin even in fully-connected graphs. 
  
\end{abstract}

\section{Introduction}\label{sec:introduction}
Bayesian inference provides a general mathematical framework for many learning tasks such as classification, denoising, object detection, and signal detection. The wide applications include but not limited to imaging processing \cite{zhang2013denoise}, multi-input-multi-output (MIMO) signal detection in digital communication \cite{cespedes2014ep,jeon2015optimality}, inference on structured lattice \cite{10.2307/25651244}, machine learning  \cite{2018arXiv180607066M, Lin:2015:DLM:2969239.2969280, yoon2019inferenceGraph}.
Specifically, statistic properties of a hidden variable $\bm{x} = \{x_1,\dots,x_N\}$ are of common interests in Bayesian inference. Practical interests usually include finding joint probability $p(\bm{x})$, marginal probability $p_i(x_i)$, the most probable state $\argmax_{\bm{x}} p(\bm{x})$. It can be extended to maximum a posterior (MAP) inference when it is conditional on some observation ($\argmax_{\bm{x}}p(\bm{x} | \cdot)$). Direct inference from $p(\bm{x})$ may be difficult computationally or technically. For instance, in the MAP inference problem, it could be the case that the gradient or subgradient of $p(\bm{x})$ may not exist and it is computationally prohibitive to search $\bm{x}$'s whole feasible space.

Probabilistic graphical models as structured graphs provide a framework for modeling the dependency between random variables. Belief propagation (BP) is a general message-passing algorithm for performing inference on graphical models. The intuition of BP is exchange of belief (statistical information) between neighboring nodes \cite{Bishop:2006:PRM:1162264}. When belief exchange converges, inference can be done by using the converged belief in graphical models. BP can solve inference problems exactly when the graphical model representation of $p(\bm{x})$ is loop-free or tree-structured \cite{kschischang2001factor_graph}.
When there are loops or circles in graphical models, BP is still a practical method to do inference approximately (loopy BP) by running it as if there is no loop. But its performance could be deteriorated significantly. In the loopy case,
there are attempts to study convergence properties of BP in special cases \cite{Ihler:2005:LBP:1046920.1088703, du2017convergenceBP}, but (loopy) BP may not converge in general.

Apart from the practical performance issues of BP in loopy graphs, the understanding of it is also limited. \cite{Yedidia:2000:GBP:3008751.3008848} shows that BP in loopy graphs approaches to a stationary point of an approximate free energy, the Bethe free energy in statistical physics. Based on this understanding, variants of BP are derived to improve BP. For instance, fractional BP in \cite{Wiegerinck:2002:FBP:2968618.2968673} applies a correction coefficient to each factor, generalized BP \cite{Yedidia:2000:GBP:3008751.3008848} propagates belief between different regions of a graph, and damping BP in \cite{Pretti2005damping} updates belief by combining old and new belief. Another track is expectation propagation (EP), introduced by Opper and Winther \cite{Opper:2000:GPC:1121900.1121911} and Minka \cite{Minka:2001:EPA:647235.720257, Minka:2001:FAA:935427}. In EP, a simpler factorized distribution defined in exponential distribution family is used to approximate the original complex distribution, and an intuitive factor-wise refinement procedure is used to find such an approximate distribution. The method has an intuition of minimizing a localized Kullback-Leibler (KL) divergence. This is discussed further in \cite{divergence-measures-and-message-passing} and it shows an unifying view of message passing algorithms. Following work stochastic EP \cite{yingzhen2015sep} explores its variant method for applications to large dataset.

In this work, we take the path of Minka's variational methods to improve BP and also to gain better understanding of BP in loopy graphs.
We define a surrogate distribution $q(\bm{x})$ first. $q(\bm{x})$ is assumed to be fully factorized and each factor of $q(\bm{x})$ represents a message in the factor graph representation of the original distribution $p(\bm{x})$. Fully factorization is the only requirement to $q(\bm{x})$. Then we define a message passing rule that is derived by minimizing a localized $\alpha$-divergence. This is factor-wise refinement of $q(\bm{x})$ iteratively. We refer to the obtained algorithm by $\alpha$-BP. The merits of $\alpha$ are as follows:
\begin{itemize}
\item[a.]{$\alpha$-BP has clear intuition as localized minimization of $\alpha$-divergence between original distribution $p$ and surrogate distribution $q$.}
\item[b.]{$\alpha$-BP generalizes the standard BP, since the message rule of BP is a special case of $\alpha$-BP.}
\item[c.]{$\alpha$}-BP could outperform BP significantly even in full-connected graphs while still maintaining simplicity of BP for inference. 
\end{itemize}

\section{Preliminary}\label{sec:preliminary}
In this section, we provide the preliminaries that are needed in this paper. We introduce the $\alpha$-divergence and a graphical model that we are going to use to explain $\alpha$-BP.

\subsection{Divergence Measures}
As explained in Section~\ref{sec:introduction}, we are going to minimize $\alpha$-divergence between $p$ and $q$, which is defined as follows according to \cite{Zhu95informationgeometric}\cite{divergence-measures-and-message-passing}: \\
\begin{equation}\label{eq:alpha-divergence}
  \Dd_{\alpha}(p \| q ) = \frac{\int_{\bm{x}} \alpha p(\bm{x}) + (1-\alpha) q (\bm{x}) - p(\bm{x})^{\alpha} q(\bm{x})^{1-\alpha} d\bm{x}}{\alpha(1-\alpha)},
\end{equation}
where $\alpha$ is the parameter of $\alpha$-divergence, distribution $p$ and $q$ are unnormalized, i.e. $\int_{\bm{x}}p(\bm{x}) d\bm{x} \neq 1$, $\int_{\bm{x}}q(\bm{x}) d\bm{x} \neq 1$.

The classic KL divergence is defined as
\begin{equation}
  KL(p \| q) = \int p(\bm{x}) \log{\frac{p(\bm{x})}{q(\bm{x})}} d \bm{x}+ \int q(\bm{x}) - p(\bm{x}) d\bm{x}
\end{equation}
where the $\int q(\bm{x}) - p(\bm{x}) d\bm{x}$ is a correction factor to accommodate unnormalized $p$ and $q$. The KL divergence is a special case of $\alpha$-divergence, since $\lim_{\alpha \rightarrow 1}\Dd_{\alpha}(p \| q ) = KL(p\|q)$ and $\lim_{\alpha \rightarrow 0}\Dd_{\alpha}(p \| q ) = KL(q\|p)$, by applying L'H\^opital's rule to \autoref{eq:alpha-divergence}.

Both $\alpha$-divergence and KL divergence are equal to zero if $p=q$, and they are non-negative (therefore satisfy the basic property of error measure).
Denote KL-projection by
\begin{equation}
  \text{proj}[p] = \uargmin{q \in \Ff} KL(p\|q),
\end{equation}
where $\Ff$ is the distribution family of $q$.

According to the stationary point equivalence Theorem in \cite{divergence-measures-and-message-passing}, $\text{proj}[p^{\alpha}q^{1- \alpha}]$ and $\Dd_{\alpha}(p\|q)$ have same stationary points. A heuristic scheme to find $q$ minimizing $\Dd_{\alpha}(p\|q)$ is to find its stationary point by a fixed-point iteration:
\begin{equation}\label{eq:fixed-point-iter}
  q(\bm{x})^{\text{new}}  = \text{proj}[p(\bm{x})^{\alpha}q(\bm{x})^{1-\alpha}].
\end{equation}

\subsection{A Graphic Model}

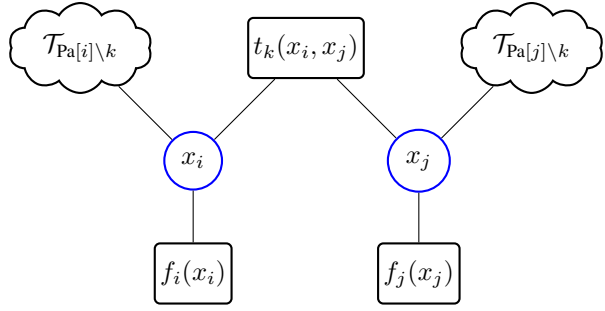
\begin{figure}
  \begin{centering}
    \begin{tikzpicture}
      \tikzstyle{enode} = [thick, draw=blue, circle, inner sep = 4pt,  align=center]
      \tikzstyle{nnode} = [thick, rectangle, rounded corners = 2pt,minimum size = 0.8cm,draw,inner sep = 2pt]

      \tikzstyle{cnode} = [thick, cloud, draw,cloud puffs=10, cloud puff arc=120, aspect=2, inner ysep=4pt]

      \node[cnode] (pajk) at (3, 1.5) {$\Tt_{\text{Pa}[j]\backslash k}$};
      \node[cnode] (paik) at (-3, 1.5) {$\Tt_{\text{Pa}[i]\backslash k}$};

      \node[nnode] (tk) at (0, 1.5) {$t_k(x_i, x_j)$};
      \node[enode] (xi) at (-1.5 ,0) {$x_i$};
      \node[nnode] (fi) at (-1.5 , -1.5) {$f_i(x_i)$};

      \node[enode] (xj) at (1.5 ,0) {$x_j$};
      \node[nnode] (fj) at (1.5 , -1.5) {$f_j(x_j)$};

      \draw[-] (xi) to (fi);
      \draw[-] (xi) to (tk);
      \draw[-] (xi) to (paik);

      \draw[-] (xj) to (fj);
      \draw[-] (xj) to (tk);
      \draw[-] (xj) to (pajk);
    \end{tikzpicture}
    \caption{Factor graph illustration of \autoref{eq:mrf}.}\label{fig:factor-graph}
    \vspace{0.1cm}
  \end{centering}
\end{figure}
We introduce a pairwise Markov random field (MRF) $p(\bm{x})$ to explain our algorithm. Variable $\bm{x} \in \Aa^N$, where $\Aa$ is a discrete finite set or subset of $\RR$ and $N$ is a positive integer. We factorize the distribution $p(\bm{x})$ as
\begin{equation}\label{eq:mrf}
  p(\bm{x}) \propto \prod_{{i=1}}^{N} f_i(x_i) \prod_{k \in \Kk} t_k(x_i, x_j),
\end{equation}
where $f_i$ is the \textit{singleton factor}, $t_k$ is \textit{pairwise factor}, $\Kk$ is the index set of all pairwise factors, and $\propto$ denotes the fact that the only difference between two sides of $\propto$ is a constant factor.

The factor graph representing \autoref{eq:mrf} is shown in \autoref{fig:factor-graph}. In the figure, $\text{Pa}[i]$ is the index set of pairwise factors connecting to variable node $x_i$, i.e. $\text{Pa}[i]$ is subset of $\Kk$, $\backslash$ denotes exclusion. $\Tt_{\text{Pa}[i]\backslash k}$ is the product of all pairwise factors connecting to $x_i$ except for $t_k$:
\begin{equation}
  \Tt_{\text{Pa}[i]\backslash k} = \prod_{n \in \text{Pa}[i]\backslash k}t_n.
\end{equation}



\section{$\alpha$-BP as Fully-Factorized Approximation}
In this section, we will show why $\alpha$-BP as a message-passing algorithm can be used as a fully-factorized approximation to the original distribution $p(\bm{x})$.

\subsection{$\alpha$ Belief Propagation}
\subsubsection{Fully Factorized Surrogate}
Now we formulate a surrogate distribution as
\begin{equation}
  q(\bm{x}) \propto \prod_{{i=1}}^{N} \tilde{f}_i(x_i) \prod_{k\in \Kk} \tilde{t}_k(x_i, x_j), \bm{x} \in \Aa^N
\end{equation}
to approximate $p(\bm{x})$. The surrogate distribution would be used to estimate inference problems of $p(\bm{x})$. We further assume that $q(\bm{x})$ can be fully factorized, which means that $\tilde{t}_k(x_i, x_j)$ can be factorized as two independent functions of $x_i, x_j$ respectively. We denote this factorization as
\begin{equation}
  \tilde{t}_k(x_i, x_j) = m_{k\rightarrow i}(x_i) m_{k\rightarrow j}(x_j).
\end{equation}
We use the notation $m_{k\rightarrow i}(x_i)$ to denote the factor as a function of $x_i$ due to the intuitive fact that $m_{k\rightarrow i}$ is also the message from the factor $t_k(x_i, x_j)$ to variable node $x_i$. Similarly we have factor $m_{k\rightarrow j}(x_j)$. Then the marginal can be formulated straightforwardly as
\begin{equation}
  q_i(x_i) \propto \tilde{f}_i(x_i) \prod_{k\in \text{Pa}[i]} m_{k\rightarrow i}(x_i).
\end{equation}

\subsubsection{Local $\alpha$-Divergence Minimization}

Now, we are going to use the heuristic scheme as in \autoref{eq:fixed-point-iter} to minimize the information loss by using tractable $q(\bm{x})$ to represent $p(\bm{x})$. The information loss is measured by $\alpha$-divergence $\Dd_{\alpha}(p(\bm{x}) \| q(\bm{x}))$.

We do factor-wise refinement to update the factors of $q(\bm{x})$ such that $q(\bm{x})$ approaches $p(\bm{x})$ asymptotically similar to \cite{divergence-measures-and-message-passing,Minka:2001:EPA:647235.720257}. Without losing generality, we begin to refine factor $\tilde{t}_k(x_i, x_j)$. Define $q^{\backslash k}(\bm{x})$ as all other factors except for $\tilde{t}_k(x_i, x_j)$
\begin{equation}
  q^{\backslash k}(\bm{x}) = q(\bm{x})/\tilde{t}_k(x_i, x_j) \propto \prod_{{i}} \tilde{f}_i(x_i) \prod_{n\in \Kk \backslash k} \tilde{t}_n(x_i, x_j).
\end{equation}
Similarly, we have $p^{\backslash k}(\bm{x})$ as all other factors except for $t_k(x_i, x_j)$. Assume that we already have had $q^{\backslash k}(\bm{x})$ as a good approximation of $p^{\backslash k}(\bm{x})$, i.e. $q^{\backslash k}(\bm{x}) \simeq p^{\backslash k}(\bm{x})$, it is $\tilde{t}_k(x_i, x_j)$ that remains to be refined. 
Then the problem $\uargmin{\tilde{t}_k^{\text{new}}} \Dd_{\alpha}\left(  p^{\backslash k}t_k\|q^{\backslash k}\tilde{t}_k^{\text{new}}\right)$ becomes \vspace{-0.3cm}
\begin{equation}
  \uargmin{\tilde{t}_k^{\text{new}}(x_i, x_j)} \Dd_{\alpha}\left(  q^{\backslash k}(\bm{x}) t_k(x_i, x_j)\|q^{\backslash k}(\bm{x}) \tilde{t}_k^{\text{new}}(x_i, x_j) \right),
\end{equation}
which searches for new factor $\tilde{t}_k^{\text{new}}$ such the above divergence is minimized.
Using \autoref{eq:fixed-point-iter}, the above problem is equivalent to
\begin{align}\label{eq:update-rule}
  &q^{\backslash k}(\bm{x}) \tilde{t}_k^{\text{new}}(x_i, x_j) \nonumber\\
  &\propto \text{proj}\left[ \left(q^{\backslash k}(\bm{x}) t_k(x_i, x_j)  \right)^{\alpha} \left(q^{\backslash k}(\bm{x}) \tilde{t}_k(x_i, x_j)  \right)^{1-\alpha} \right] \nonumber \\
  & \propto \text{proj}\left[ q^{\backslash k}(\bm{x}) t_k(x_i, x_j)^{\alpha} \tilde{t}_k(x_i, x_j)^{1-\alpha} \right].
\end{align}

Let us refine one message per time in factor $\tilde{t}_k$. Without lose of generality, we update $m_{k\rightarrow i}$ and denote
\begin{equation}
  \tilde{t}_k^{\text{new}}(x_i, x_j) = m_{k\rightarrow i}^{\text{new}}(x_i) m_{k\rightarrow j}(x_j).
\end{equation}
Since KL-projection to a fully factorized distribution reduces to matching the marginals, \autoref{eq:update-rule} is reduced to
\begin{equation}\label{eq:message-update}
  \sum_{\bm{x}\backslash x_i} q^{\backslash k}(\bm{x}) \tilde{t}_k^{\text{new}}(x_i, x_j) \propto \sum_{\bm{x}\backslash x_i} q^{\backslash k}(\bm{x}) t_k(x_i, x_j)^{\alpha} \tilde{t}_k(x_i, x_j)^{1-\alpha}.
\end{equation}
We use summation here. But it should be replaced by integral if $\Aa$ is a continuous set.
Solving \autoref{eq:message-update} gives the message passing rule as
\begin{align}\label{eq:message-rule}
  {m}^{\text{new}}_{k\rightarrow i}(x_i) \propto & \bigg[ \sum_{x_j} t_k(x_i, x_j)^{\alpha} {m}_{k\rightarrow j}(x_j)^{1-\alpha} m_{j \rightarrow k}(x_j) \bigg] \nonumber \\
                                                 &\cdot m_{k\rightarrow i}(x_i)^{1-\alpha},
\end{align}
where
\begin{equation}
  m_{j \rightarrow k}(x_j) = \tilde{f}_j(x_j) \prod_{n \in \text{Pa}[j] \backslash k} m_{n \rightarrow j}(x_j).
\end{equation}
Similarly, the message from $t_k$ to $x_j$, $m_{k \rightarrow j}(x_j)$, can be updated in similar way.

As for the singleton factor $\tilde{f}_i(x_i)$, we can do the refinement procedure on $\tilde{f}_i(x_i)$ in the same way as we have done for $\tilde{t}_k(x_i, x_j)$. This gives us the update rule of $\tilde{f}_i(x_i)$ as
\begin{equation}\label{eq:fix-factor-update}
  \tilde{f}_i^{\text{new}}(x_i) \propto f_i(x_i)^{\alpha} \tilde{f}_i(x_i)^{1-\alpha},
\end{equation}
which is the belief from factor $f_i(x_i)$ to variable $x_i$. Note, if we initialize $\tilde{f}_i(x_i) = f_i(x_i)$, then it remains the same in all iterations.

\subsection{Remarks on $\alpha$-BP}\label{subsec:remark}
\begin{figure}[!ht]
  \begin{centering}
    \begin{tikzpicture}
      \tikzstyle{enode} = [thick, draw=blue, circle, inner sep = 4pt,  align=center]
      \tikzstyle{nnode} = [thick, rectangle, rounded corners = 2pt,minimum size = 0.8cm,draw,inner sep = 2pt]

      \tikzstyle{cnode} = [thick, cloud, draw,cloud puffs=10, cloud puff arc=120, aspect=2, inner ysep=4pt]

      \node[cnode] (paik) at (0, 1.5) {$\Tt_{\text{Pa}[i]}$};

      \node[enode] (xi) at (0 ,0) {$x_i$};
      \node[nnode] (fi) at (0 , -1.5) {$f_i(x_i)$};
      \node[nnode] (pi) at (2, 0) {$\hat{p}_i(x_i)$};
      \draw[-] (xi) to (fi);
      \draw[-] (xi) to (pi);
      \draw[-] (xi) to (paik);
    \end{tikzpicture}
    \caption{Factor graph illustration with prior factor.}\label{fig:factor-graph-with-prior}
    \vspace{0.1cm}
  \end{centering}
\end{figure}
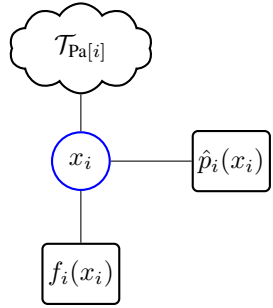

As discussed in Section~\ref{sec:preliminary}, $KL(p\|q)$ is the special case of $\Dd_{\alpha}(p\|q)$ when $\alpha \rightarrow 1$. When applying $\alpha=1$ to \autoref{eq:message-rule}, it gives
\begin{equation}
  {m}^{\text{new}}_{k\rightarrow i}(x_i) \propto \sum_{x_j} t_k(x_i, x_j) m_{j \rightarrow k}(x_j),
\end{equation}
which is exactly the messages of BP in Chapter~$8$ of \cite{Bishop:2006:PRM:1162264}. From this point of view, $\alpha$-BP generalizes BP.

Inspired by \cite{pseudo_priorBP2010} and assembling methods \cite{James:2014:ISL:2517747}, we can add an extra singleton factor to each $x_i$ as prior information that is obtained from other (usually weak) methods. This factor stands for our belief from exterior estimation. Then run our $\alpha$-BP. Denote the prior by $\hat{p}_i(x_i)$ for variable node $x_i$, then the factor graph including this prior belief can be represented as in \autoref{fig:factor-graph-with-prior}.

We summarize the method into the pseudo-code in Algorithm~\autoref{alg:alphaBP}. Though we explain the method with a binary MRF, it is straightforward to replace the factor $t_k$ by a factor involving more than two variables and applies $\alpha$-BP to general factor graphs.
\begin{algorithm}
  \caption{Algorithm of $\alpha$-BP}\label{alg:alphaBP}
  \begin{algorithmic}[1]
    \renewcommand{\algorithmicrequire}{\textbf{Input:}}
    \renewcommand{\algorithmicensure}{\textbf{Output:}}
    \REQUIRE Factor graph of $p(\bm{x})$
    \STATE Initialize $q(\bm{x})$
    \IF {Prior belief on $x_i$ available}
    \STATE Add prior factor as \autoref{fig:factor-graph-with-prior}
    \ENDIF
    \WHILE{not converge}
    \FOR {each edge of factor graph}
    \STATE Message update by \autoref{eq:message-rule} or \autoref{eq:fix-factor-update}
    \ENDFOR
    \ENDWHILE
    \RETURN $q(\bm{x})$ 
  \end{algorithmic} 
\end{algorithm}

\section{Experimental Results}
In this section, we report numerical results on the $\alpha$-BP. It is well known that performance of BP and its variants deteriorate significantly when loops appear in factor graph. We would like to see if $\alpha$-BP could relief the deterioration brought by loops in inference. Thus we firstly test the performance of $\alpha$-BP for MAP inference in a MRF with $\Aa=\{-1,1\}$ (Ising model), where we adjust how loopy its corresponding factor graph is.

In addition, we apply the $\alpha$-BP to a MIMO detection problem, to explore its performance in comparison with (loopy) BP and minimum mean square error (MMSE). At the end, the prior factor trick is used according to discussion in Subsection~\ref{subsec:remark}. This turns out to be MAP inference problem as well.

For the MAP inference, the most probable estimation by $\alpha$-BP, $\hat{\bm{x}}=[\hat{x}_1, \cdots, \hat{x}_N]$, is obtained by
\begin{align}
  \hat{x}_{i} = \uargmax{x_i}\tilde{f}_i(x_i) \prod_{k\in \text{Pa}[i]} m_{k\rightarrow i}(x_i), x_i \in \Aa.
\end{align}

\subsection{$\alpha$-BP on binary pairwise MRF}
\begin{figure*}[!ht]
  \begin{subfigure}{.339\textwidth}
    \captionsetup[subfigure]{justification=centering}
    \centering
    \includegraphics[width=1\linewidth]{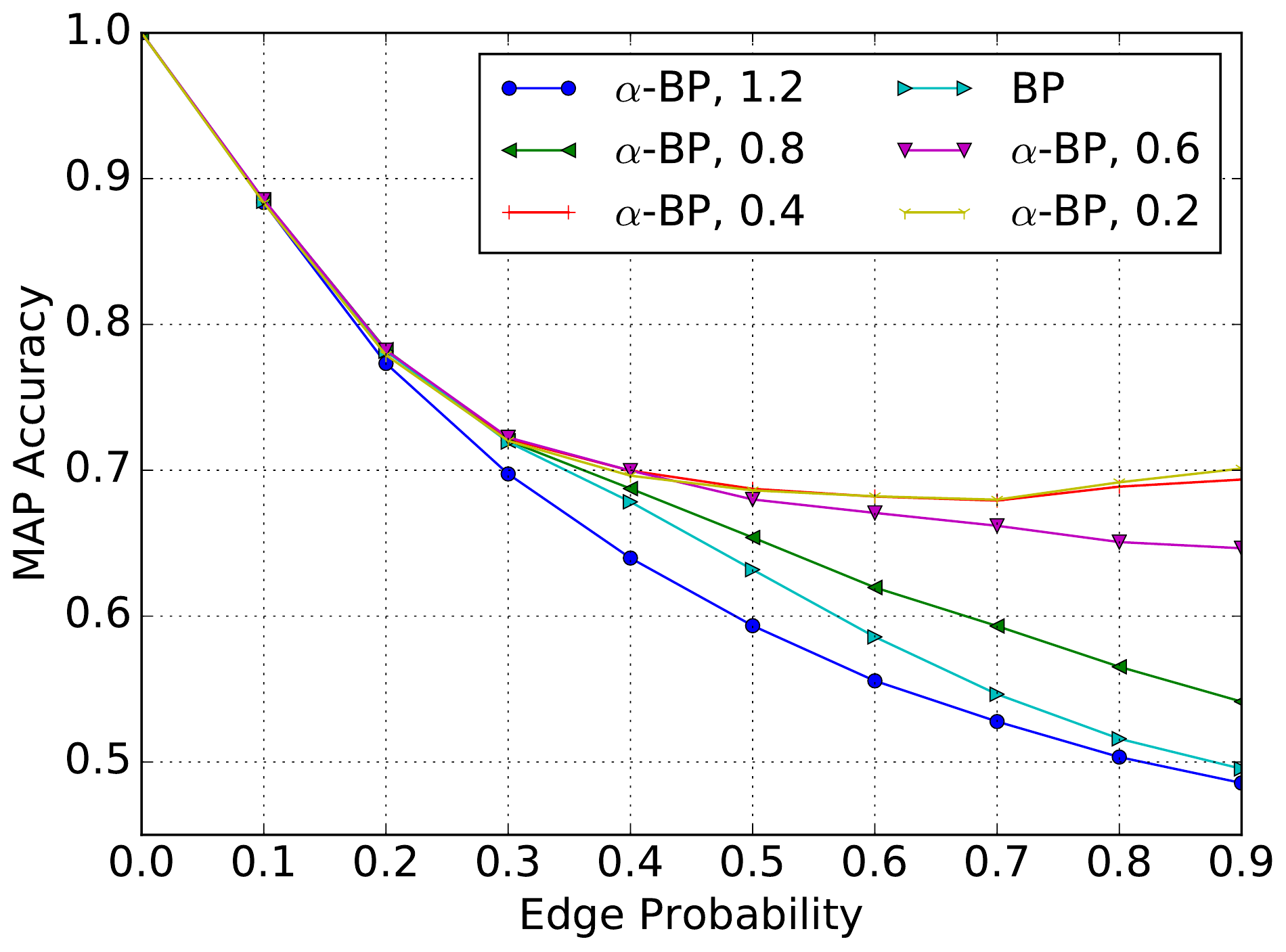}
    \vspace{-0.6cm}
    \caption{Mismatch between MAP and $\alpha$-BP on\\ binary MRF}
    \label{fig:mismatch}
  \end{subfigure}
  \begin{subfigure}{.33\textwidth}
    \includegraphics[width=1\linewidth]{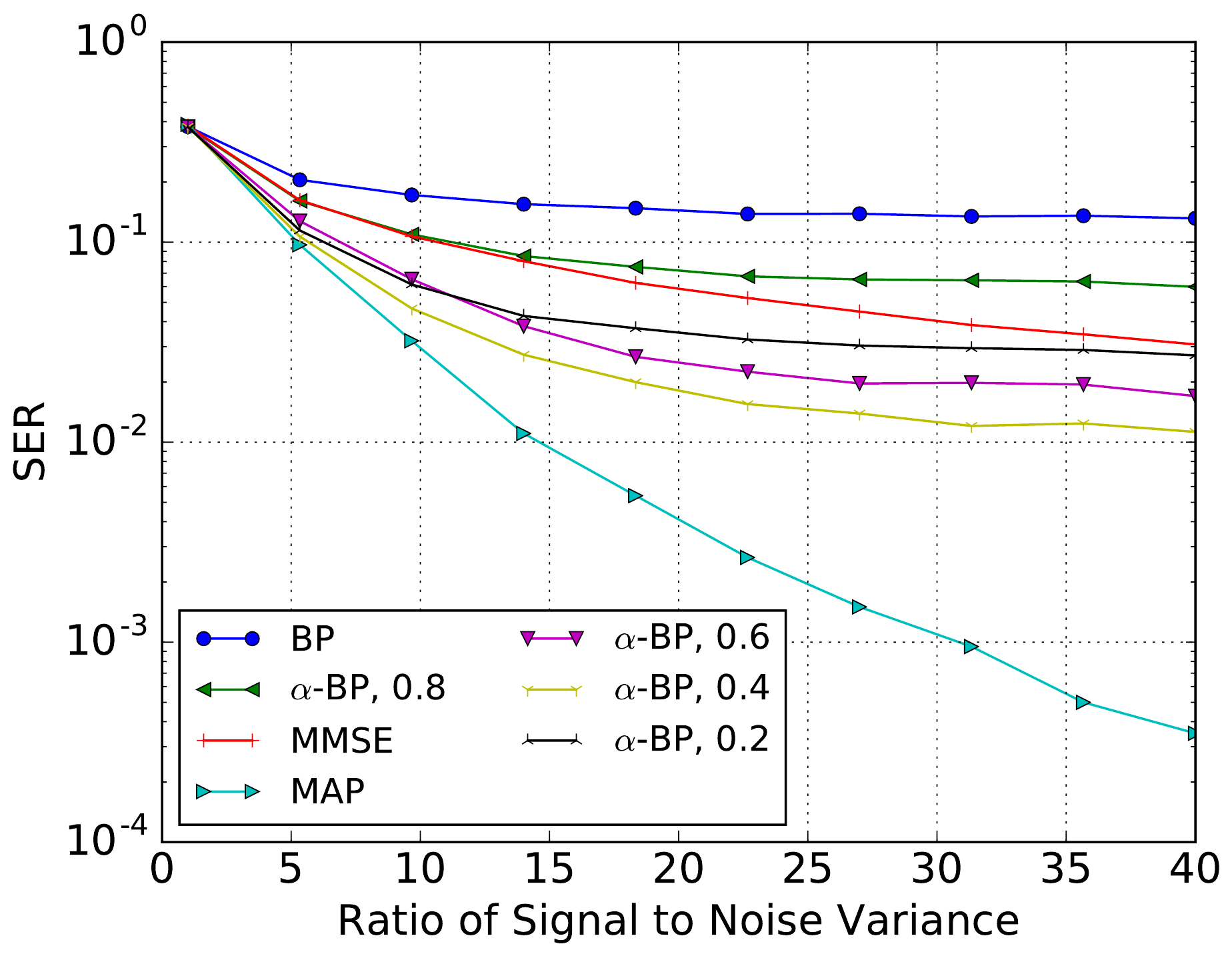}
    \vspace{-0.6cm}
    \caption{MIMO detection: $\alpha$-BP without prior\\~}\label{fig:mimo_a}
  \end{subfigure}
  \begin{subfigure}{.33\textwidth}
    \includegraphics[width=1\linewidth]{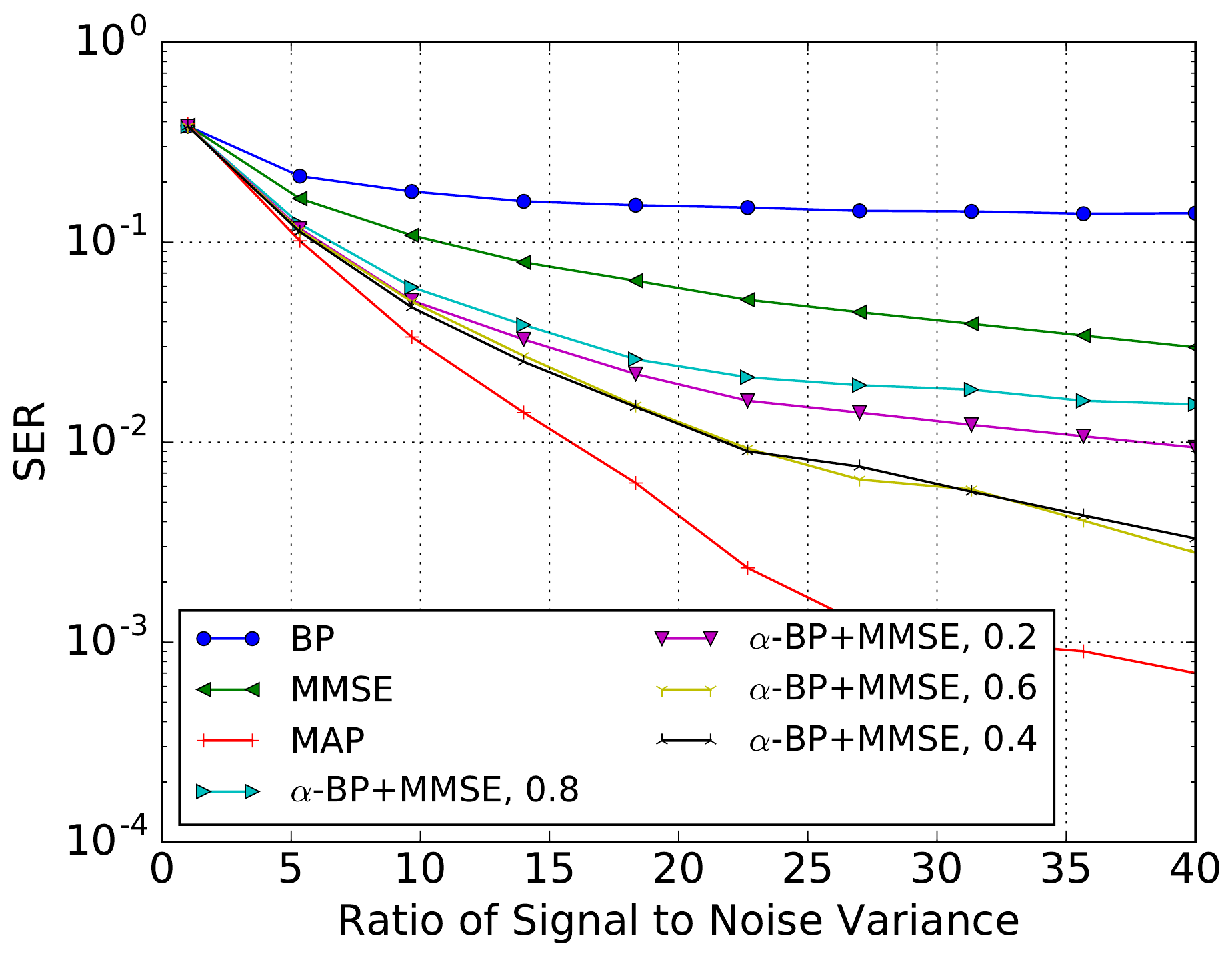}
    \vspace{-0.6cm}
    \caption{MIMO detection: $\alpha$-BP with prior\\~}\label{fig:mimo_b}
  \end{subfigure}
  \vspace{-0.3cm}
  \caption{Numerical Results of $\alpha$-BP on: (a) binary MRF, (b) and (c) MIMO detection.}
  \vspace{0.3cm}
  \label{fig:mimo_detection}
\end{figure*}
With $t_k(x_i, x_j) = e^{-2J_{i,j}x_i x_j}$ and $f_i(x_i) = e^{-J_{i,i}-b_i x_i}$, \autoref{eq:mrf} can be reformulated as
\begin{equation}
  p(\bm{x}) \propto \exp\{-\bm{x}^{T}\bm{J}\bm{x} - \bm{b}^{T}\bm{x}\}, \bm{x} \in \Aa^N,
\end{equation}
where $\bm{x}^{T}$ is transform of $\bm{x}$, $J_{i,j}$ is element of symmetrix matrix $\bm{J}$ at $i$-th row and $j$-th column, $\bm{b} = [b_1, \cdots, b_N]^T$.

For this experiment, we set $\Aa = \left\{ -1, 1 \right\}$ and $N=9$. Bias $\bm{b}$ is sampled from Gaussian, ${b_i} \sim \Nn(0, (1/4)^2)$. Since $\bm{J}$ decides the loopy level of its corresponding factor graph, we use the Erdos-Rényi model \cite{erdos1960} to construct its connectivity. Namely, an element of $\bm{J}$ is set as non-zero, $ {J}_{i,j} = {J}_{j,i} \sim \Nn(0,1)$, with an \textit{Edge Probability}. Otherwise, ${J}_{i, j} = {J}_{j,i} = 0$, which means this is no connection between variable node $x_i$ and $x_j$. For each test value of Edge Probability, $5000$ binary MRF models are generated randomly and mismatch between $\argmax_{\bm{x}} p(\bm{x})$ and $\left\{ \bm{x}|\argmax_{x_i}q_i(x_i) \right\}$ is computed in each realization. The results are shown in \autoref{fig:mismatch}. As the Edge Probability increases, graphs become loopier and $\alpha$-BP (also BP) has more mismatch with MAP inference. In general, $\alpha$-BP with $\alpha > 1$ underperforms BP, and $\alpha$-BP with $\alpha < 1$ outperforms BP. For $\alpha=0.2, 0.4, 0.6$, $\alpha$-BP stops deteriorating for Edge Probability increasing over $0.35$, while BP continues giving even worse approximations.

\subsection{Application to MIMO Detection}

For MIMO system, the observation $\bm{y}$ is a linear function of channel $\bm{H}\in \RR^{M\times N}$ when unknown signal $\bm{x}$ need to be estimated:
\begin{equation}\label{eq:linear-model}
  \bm{y} = \bm{H} \bm{x} + \bm{e}, \bm{x} \in \Aa^N,
\end{equation}
where $\bm{e}$ is  noise modeled as Gaussian noise $ \bm{e} \sim \Nn\left( \bm{0}, \sigma^2_{w} \bm{I} \right)$. Here $\bm{I}$ is unitary matrix. In this case, the posterior of $\bm{x}$ can be written as:
\begin{align}\label{eq:true-posterior}
  &p(\bm{x}|\bm{y}) \propto \exp\left\{ - \frac{1}{2\sigma_w^2} \norm{\bm{Hx} - \bm{y}}^2 \right\} \nonumber \\
  & = \exp\left\{ - \frac{1}{2\sigma_w^2}\left[ \bm{x}^{T}\bm{H}^{T}\bm{H}\bm{x} - 2 \bm{y}^T\bm{H}\bm{x}  + \bm{y^T}\bm{y}  \right] \right\}
\end{align}
where $\norm{\cdot}$ is the Euclidean norm. Denote $\bm{S} = \bm{H}^T\bm{H}$, $\bm{h}_i$ as the $i$-th column of $\bm{H}$, and apply
\begin{align}
  f_{{i}}(x_i) &= \exp\left\{- \frac{S_{i,i} x_i^2}{2 \sigma_w^2} + \frac{\langle {\bm{h}_i, \bm{y}}\rangle x_i}{\sigma_w^2} \right\}, \\
  t_{{k}}(x_i, x_j) &= \exp\left\{ -\frac{x_i S_{i,j} x_j}{\sigma_w^2} \right\}.
\end{align}
Then the \autoref{eq:true-posterior} is equivalent to \autoref{eq:mrf}. We set $\Aa = \left\{ -1, 1 \right\}$, $N = 8$, and $\bm{H}\in \RR^{8 \times 8}$ sampled from Gaussian.

We test the application of $\alpha$-BP to the MIMO signal detection numerically. We run the $\alpha$-BP, without prior trick in \autoref{fig:mimo_a} and with prior in \autoref{fig:mimo_b} (legend ``$\alpha$-BP$+$MMSE'') from MMSE. The reference results of MMSE and MAP inference are also reported under the same conditions. MMSE estimator depends on Gaussian posterior $\Nn(\hat{\bm{\mu}}, \hat{\bm{\Sigma}})$ with $\hat{\bm{\mu}} = (\bm{H}^{T}\bm{H} + \sigma_w^2 \bm{I})^{-1}\bm{H}^{T}\bm{y}$ and $\hat{\bm{\Sigma}} = (\bm{H}^{T}\bm{H} + \sigma_w^2 \bm{I})^{-1}\sigma_w$. Detection of MMSE is carried out by $\argmin_{x_i\in \Aa}\abs{x_i-\hat{\mu}_i}$.

\autoref{fig:mimo_a} shows that BP even underperforms MMSE but $\alpha$-BP can outperform MMSE by assigning smaller value of $\alpha$.
Note that MMSE requires the matrix inverse computation whose complexity is proportional to $N^3$, while the complexity of $\alpha$-BP increases linearly with $N$. Therefore $\alpha$-BP is superior to MMSE both performance-wise and complexity-wise.  
However, there is still a big gap between $\alpha$-BP (even for $\alpha=0.4$) and MAP. This gap can be decreased further by using the prior trick discussed in Subsection~\ref{subsec:remark}. \autoref{fig:mimo_b} exemplifies this effects by using prior belief from MMSE, $\hat{p}_i(x_i)\propto \exp\{-(x_i-\mu_i)^2/(2\bm{\Sigma}_{i,i})\}$, which comes with legend "$\alpha$-BP$+$MMSE". It is shown that larger performance gain is observed when $\alpha$-BP runs with prior belief.

\section{Conclusion}
In this work, we obtain the $\alpha$-BP method by doing local $\alpha$-divergence minimization between model distribution $p$ and surrogate distribution $q$. $\alpha$-BP is a practical Bayesian method for message passing. $\alpha$-BP is a valid and practical algorithm accoinding to our experiments. With prior trick, the performance of $\alpha$-BP can be further improved. Future works would be to investigate the parallel message passing scheduling of $\alpha$-BP. It is also interesting to study guideline on choice of $\alpha$.

\bibliographystyle{IEEEtran}

\bibliography{myref}


\end{document}